%% file: root.tex
\documentclass[letterpaper, 10 pt, conference]{ieeeconf}  
\usepackage{graphicx}

\IEEEoverridecommandlockouts                              

\usepackage{amsmath} 
\usepackage{amssymb}  
\usepackage{algorithm}
\usepackage{algpseudocode}
\usepackage{booktabs}
\usepackage{makecell}
\usepackage{amssymb}
\usepackage[normalem]{ulem}

\makeatletter
\renewcommand\fs@ruled{%
  \def\@fs@cfont{\bfseries}%
  \let\@fs@capt\floatc@ruled
  \def\@fs@pre{\kern5pt\hrule height.8pt depth0pt \kern2pt}%
  \def\@fs@post{\kern2pt\hrule\relax}%
  \def\@fs@mid{\kern2pt\hrule\kern2pt}%
  \let\@fs@iftopcapt\iftrue
}
\makeatother

\newif\ifmarkup \markupfalse
\DeclareRobustCommand{\revadd}[1]{\ifmarkup\textbf{#1}\else#1\fi}
\DeclareRobustCommand{\revdel}[1]{\ifmarkup\sout{#1}\fi}

\title{\LARGE \bf
Contact-Rich Motion Planning via GPU-Parallel Mode Evaluation 
}

\author{Jiayun Li$^{1,2}$ and Georgia Chalvatzaki$^{1,2,3}$%
\thanks{$^{1}$PEARL Lab, Department of Computer Science, TU Darmstadt, Darmstadt, Germany.}%
\thanks{$^{2}$Hessian.AI, Darmstadt, Germany.}%
\thanks{$^{3}$Robotics Institute Germany (RIG), Germany.}%
}

\begin{document}

\maketitle
\thispagestyle{empty}
\pagestyle{empty}

\begin{abstract}
Contact-rich motion planning (CRMP) is essential for robotic manipulation and locomotion, yet remains computationally challenging due to combinatorial contact decisions. Existing methods typically avoid broad evaluation of contact-mode sequences through search heuristics or optimization reformulations. We revisit broad evaluation in light of modern GPU hardware and introduce Contact-Mode Expansion with parallel Trajectory optimization (CoMET), which combines GPU-parallel trajectory evaluation with greedy contact-mode expansion. On planar pushing benchmarks, CoMET is competitive with optimization-based, sampling, and tree-search baselines in solution quality and planning time, matching the full-enumeration reference on nearly all instances with fewer evaluations and shorter planning times. Ablations suggest that much of the performance gain comes from the high-throughput mode evaluator. In bimanual nonprehensile manipulation, GPU-friendly local mode expansion achieves higher planning success than the tested adaptive tree search as the mode space grows. These results demonstrate that broad explicit mode evaluation provides a simple yet effective alternative for CRMP.
\end{abstract}

\input{intro}
\input{related}
\input{method}
\input{exp}

\section{CONCLUSIONS}
\label{sec:conclusion}
This paper revisited explicit contact-mode evaluation for CRMP under
modern GPU parallelism. CoMET combines a high-throughput GPU evaluator
for heterogeneous mode-conditioned constrained TO with GPU-friendly
local mode expansion. On planar pushing, CoMET was competitive with
convex-relaxation, contact-implicit, sampling, and tree-search baselines,
recovering the full-enumeration reference on nearly all instances with
substantially fewer evaluations. Ablations indicate that evaluator
efficiency and GPU execution account for much of this performance.
On bimanual nonprehensile flipping, where the mode space exceeds the
evaluation budget, local expansion under the same evaluator and budget
solves more instances than the tested adaptive tree search and uniform
sampling without degrading returned-plan quality. Real-robot experiments
further demonstrate physical feasibility. Together, these results highlight the complementary benefits of high-throughput evaluation and local mode expansion, supporting broad explicit mode evaluation as a simple yet effective option for CRMP.
\addtolength{\textheight}{-0.1cm}   







\section*{ACKNOWLEDGMENT}
This work was supported by the DFG Emmy Noether Programme (CH 2676/1-1), the EU Horizon Europe projects MANiBOT (101120823) and ARISE (101135959), the BMFTR project RIG (16ME1001), and the ERC project SIREN (101163933). We also acknowledge support from the hessian.AI Service Center (BMFTR, 16IS22091), the hessian.AI Innovation Lab (S-DIW04/0013/003), TAM, RAI, Google, and the Alfried Krupp Foundation. Claude Opus 5 (Anthropic), via Claude Code, provided programming and
debugging assistance for Sec.~III-C; all code was reviewed and validated
by the authors.



\bibliographystyle{IEEEtran}
\bibliography{reference}

\end{document}

%% file: intro.tex
\section{INTRODUCTION}
Contact-rich motion planning (CRMP) is essential for a broad class of robotic manipulation and locomotion tasks. Robots need to reason about how to make and break contact with the environment to accomplish specific tasks. This requires deciding when, where, and how strongly to create task-relevant contacts, while accounting for physical and geometrical feasibility, solution optimality, and computational efficiency. Depending on the formulation, CRMP typically leads to either nonsmooth optimization or combinatorial reasoning over contact-mode sequences, making it difficult to find reliable solutions efficiently. 

The hybrid structure of CRMP naturally gives rise to a mixed-integer nonlinear program (MINLP), in which discrete variables encode contact modes and continuous variables describe the corresponding motions and contact interactions. Classical approaches address this structure by explicitly specifying or searching over contact modes, reducing the optimization within each mode to a continuous problem \cite{lozano1984automatic,5345770,toussaint2015ijcai-logic,doshi2020hybrid, cheng2022contact}. The central difficulty is therefore generally regarded as managing the combinatorial mode-search without incurring excessive continuous optimization. Consequently, search-based methods typically use heuristics or selective expansion to avoid evaluating a large number of candidate mode sequences. More recently, Graph-of-Convex-Sets (GCS) methods have been proposed to address the combinatorial contact-mode search through convex relaxation \cite{marcucci2024shortest,Graesdal-RSS-24}. However, solving the resulting relaxations can be computationally expensive, particularly for problems with large contact-mode spaces. Alternatively, contact-implicit trajectory optimization (CITO) avoids explicit mode-sequence evaluation by jointly optimizing motion and contact forces \cite{posa2014direct,manchester2019contact,patel2019contact}. This formulation, however, leads to a highly nonconvex optimization problem that can be sensitive to initialization and hyperparameters, making reliable convergence challenging \cite{onol2020tuning}.

Recent advances in GPU-accelerated trajectory optimization (TO) have shifted the practical computational regime, making it possible to optimize many candidates in parallel \cite{ShenW-RSS-25,shen2026parallel}. Contact-mode evaluation, however, presents a heterogeneous batch: candidates differ in active contact constraints and convergence behavior. This raises a natural question: \emph{How can heterogeneous mode-conditioned TO be evaluated efficiently on GPUs, and how should the resulting evaluation budget be allocated?}

\begin{figure}[t]
    \centering
    \includegraphics[width=\columnwidth]{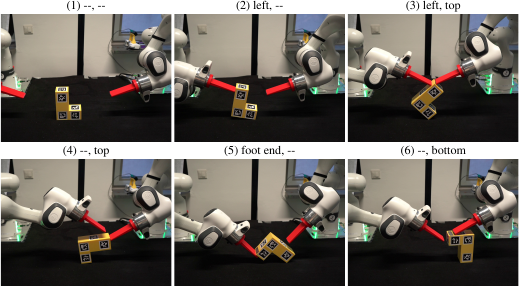}
    \caption{Real-robot bimanual nonprehensile stable L-shape object flip.
    Panels are numbered in time order; labels give the block
    face touched by the left and right paddle, named in the block's
    initial pose (panel 1); ``-{}-'' denotes free motion.}
    \label{fig:placeholder}
\end{figure}

In this work, we revisit explicit contact-mode evaluation as a planning primitive under modern GPU parallelism. We design a GPU-oriented primal-dual augmented-Lagrangian (AL) differential dynamic programming (DDP) evaluator \cite{jallet2022constrained,adeoye2025proximal} that maps mode-dependent contact constraints to a shared fixed structure and independently advances thousands of mode-conditioned TO problems. Rather than sequentially selecting and refining individual mode sequences, our method evaluates large batches and greedily expands the neighborhoods of the best sequences found so far. We call the resulting framework \textbf{Contact-Mode Expansion with parallel Trajectory optimization} (\textbf{CoMET}).

Experiments on planar pushing and bimanual non-prehensile manipulation highlight the contributions of high-throughput trajectory evaluation and local mode expansion across different mode-space sizes. On the planar pushing benchmark of \cite{Graesdal-RSS-24}, CoMET recovers the full-enumeration reference on nearly all instances while evaluating only a fraction of the admissible sequences, and returns plans up to \(4.4\times\) faster than the strongest optimization-based baseline while achieving lower objective values under a matched criterion. Controlled ablations suggest that much of this performance gain comes from the trajectory evaluator, while the tested adaptive tree-search allocation offers no observed advantage over parallel mode expansion under the same evaluator and compute budget. On the bimanual task, where the admissible space exceeds half a million sequences and only a small fraction can be evaluated, GPU-friendly mode expansion achieves higher planning success than the tested adaptive tree search and uniform sampling. Together, these results highlight the complementary benefits of high-throughput evaluation and local mode expansion. In real-robot experiments, the planned bimanual motions flip the L-shaped block through multiple support transitions with an \(80\%\) success rate. 

To summarize, our main contributions are: (1) a GPU-parallel evaluator that executes heterogeneous mode-conditioned constrained TO problems with shared fixed structure and independent candidate progress; (2) a GPU-friendly neighborhood-expansion strategy for allocating the evaluation budget; (3) an empirical comparison with convex-relaxation, contact-implicit, sampling, and Monte Carlo tree search (MCTS) baselines, along with ablation studies examining the factors contributing to performance gains; and (4) real-world experiments demonstrating the physical feasibility of the generated plans. Our implementation of CoMET will be released as open source upon acceptance.

%% file: related.tex
\section{RELATED WORK}
We position CoMET within three related lines of research: discrete reasoning over contact modes, GPU-parallel TO and planning, and reformulations of contact decisions.

\subsection{Discrete Reasoning over Contact Modes}

A common strategy for CRMP is to search over contact-mode sequences while solving continuous motion optimization conditioned on each sequence. Tree-search methods selectively expand promising mode sequences using feasibility or cost information \cite{chen2021trajectotree,cheng2022contact,natarajan2023torque}, while Logic-Geometric Programming uses hierarchical bounds and geometric optimization to reduce expensive nonlinear solves \cite{toussaint2015ijcai-logic,toussaint2017multi,Toussaint-RSS-18,ortiz2022conflict}. Sampling-based methods similarly concentrate computation on selected contact sequences \cite{cheng2022contact,Zhang-RSS-23}. CoMET instead evaluates large batches of mode sequences with GPU-parallel constrained TO, shifting the emphasis from avoiding evaluations to performing them efficiently in parallel.

\subsection{GPU-Parallel Trajectory Optimization and Planning}

GPU acceleration has enabled high-throughput TO through parallel dynamics, numerical optimization, and multiple candidate trajectories \cite{le2025global,sundaralingam2023curobo,adabag2024mpcgpu,du2025gato}. At the task planning level, cuTAMP evaluates thousands of continuous candidates with GPU-parallel differentiable optimization \cite{ShenW-RSS-25}, but does not solve contact-constrained, dynamics-coupled trajectories within each candidate. CoMET instead uses GPU parallelism to solve large batches of contact-mode-conditioned constrained TO problems directly.

\subsection{Reformulations of Contact Decisions}

Contact-implicit TO avoids prescribed mode sequences by allowing contacts to emerge through complementarity-based formulations \cite{posa2014direct,manchester2019contact,patel2019contact,sleiman2019contact}. Specialized methods improve their numerical tractability through augmented-Lagrangian optimization, linear complementarity dynamics, differentiable contact models, and structure-exploiting solvers \cite{LiJ2-RSS-26,Aydinoglu2024,huang2024adaptive,le2024fast}. These formulations nevertheless remain highly nonconvex and sensitive to initialization.

Convex reformulations provide another route. GCS-based semidefinite relaxations trade computational cost for stronger global reasoning \cite{Graesdal-RSS-24, KangS-RSS-25}, while convex-decomposition methods transform discrete geometric choices into continuous constrained optimization \cite{zhang2026sequential}. CoMET instead retains explicit contact modes and addresses the resulting combinatorial structure through parallel constrained optimization.

%% file: method.tex
\section{METHOD}
\label{sec:method}

\begin{figure}[t]
\vspace*{3pt}
    \centering
    \includegraphics[width=\columnwidth]{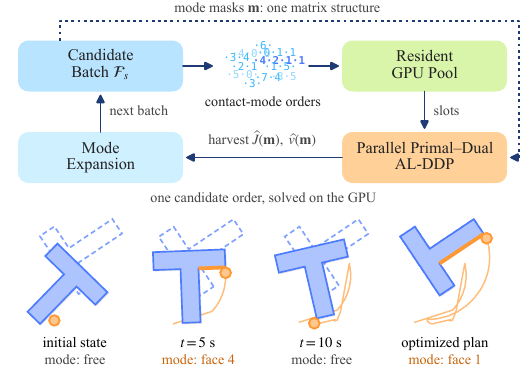}
    \caption{CoMET overview. Bottom: one candidate order on push-T
    (slider blue, pusher orange, goal dashed).}
    \label{fig:overview}
\end{figure}

CoMET combines parallel contact-mode evaluation with mode
expansion (Fig.~\ref{fig:overview}). Each candidate specifies a contact-mode order, while continuous optimization determines the trajectory, contact interactions, and transition timing. Starting from a broad seed batch, CoMET evaluates candidates in parallel under a limited optimization budget and expands the neighborhoods of orders ranked by a merit function combining objective value and constraint violation. The leading feasible candidates are then refined to obtain the final plan. We first formulate the mode-conditioned TO problem, then describe the contact mode expansion strategy and its GPU-parallel TO evaluator.

\subsection{Mode-Conditioned Trajectory Optimization}
\label{sec:problem}

We formulate CRMP as a mixed discrete--continuous optimization problem
\begin{equation}
\begin{aligned}
\min_{\mathbf m\in\mathcal S,\,z}\; J(z,\mathbf m)
\quad \mathrm{s.t.}\quad
g(z,\mathbf m)=0, \;
h(z,\mathbf m)\leq0,
\end{aligned}
\label{eq:crmp}
\end{equation}
where $z$ collects continuous trajectory variables, including states,
controls, and timing variables. The equalities $g$ encode dynamics and
contact equalities, while $h$ contains unilateral contact, friction,
non-penetration, and obstacle constraints.

The discrete decision is a contact-mode order
$\mathbf m=(m^1,\ldots,m^S)\in\mathcal S$. The admissible set
$\mathcal S$ is the language of a task-level transition system:
modes are discrete contact events (which body contacts
which face, or free motion), and the transition rules encode physical
gating, e.g., a contact changes faces only through a free-motion segment.
$\mathcal S$ contains all rule-consistent orders up to a segment
cap, with no two consecutive segments sharing a mode, so distinct
orders are not related by segment merging. The rules are specified
per task from the contact geometry alone and encode no solution
knowledge; they are common discrete structures that mode-explicit search \cite{chen2021trajectotree,cheng2022contact}
also requires.
For a fixed contact mode sequence, transition times and segment durations remain continuous variables in $z$, and Eq.~\eqref{eq:crmp} reduces to a smooth constrained TO problem,
\begin{equation}
\begin{aligned}
J^*(\mathbf m)=\min_z\; J(z,\mathbf m)
\quad \mathrm{s.t.}\quad
g(z,\mathbf m)=0, \;
h(z,\mathbf m)\leq0.
\end{aligned}
\label{eq:mode_nlp}
\end{equation}
Note that our evaluator for TO uses local optimization for each candidate mode sequence and is therefore not guaranteed to attain $J^*(\mathbf m)$.
Ideally, the contact-mode sequence would be selected as
\begin{equation}
\mathbf m^*
=
\arg\min_{\mathbf m\in\mathcal S}J^*(\mathbf m).
\label{eq:mode_selection}
\end{equation}

CoMET approximates this selection using finite-budget local solves from the TO evaluator.
For each order $\mathbf m$, let
$\widehat z=z_{t_{\mathrm{exp}}}(\mathbf m)$ denote the trajectory
returned after the exploration budget $t_\mathrm{exp}$. Its objective and constraint
violation are
\begin{equation}
\begin{aligned}
\widehat J(\mathbf m)
&= J(\widehat z,\mathbf m),\\
\widehat v(\mathbf m)
&= \max\!\left\{
\|g(\widehat z,\mathbf m)\|_\infty,\,
\|[h(\widehat z,\mathbf m)]_+\|_\infty
\right\}.
\end{aligned}
\label{eq:inexact_mode_value}
\end{equation}
A contact mode sequence is considered feasible when the continuous optimization is feasible:
$\widehat v(\mathbf m)\leq\epsilon_{\mathrm{feas}}$.
During exploration, all evaluated orders are ranked by
\begin{equation}
\widehat M(\mathbf m)
=
\widehat J(\mathbf m)
+
\mu\max\!\left\{
\widehat v(\mathbf m)-\epsilon_{\mathrm{feas}},\,0
\right\},
\label{eq:search_merit}
\end{equation}
where $\mu>0$ penalizes constraint violation beyond the feasibility
tolerance $\epsilon_{\mathrm{feas}}$. Infeasible orders can therefore guide further expansion even when no feasible solution has been found.

\subsection{Contact-Mode Expansion with Parallel TO}
\label{sec:parallel_enumeration}

CoMET uses a batch-oriented greedy expansion designed to allocate the GPU TO evaluation budget. The expansion rule is deliberately batch-synchronous: its purpose is to expose, rather than throttle, the throughput of the parallel evaluator. The seed batch $\mathcal F_0$ contains all admissible orders up to a prescribed number of segments. Let $\mathcal V$ denote all evaluated orders. After all candidates in $\mathcal F_s$ have completed, their results are added to $\mathcal V$, which is ranked by increasing merit $\widehat M$ in Eq.~\eqref{eq:search_merit}.

\textbf{Neighborhood expansion:}
An edit inserts, deletes, or substitutes one mode segment. We denote by
$\mathcal N_r(\mathbf m)$ the orders reachable from $\mathbf m$ through
at most $r$ edits, with every intermediate order remaining admissible in
$\mathcal S$. For example, a one-edit neighborhood includes alternatives obtained by changing one contact mode or adding or removing one segment. This allows large batches of candidate mode sequences to be generated around the best-ranked orders without selecting candidates one at a time, making the expansion well suited to GPU-parallel evaluation.

To construct $\mathcal F_{s+1}$, CoMET traverses $\mathcal V$ in merit
order and admits unseen orders from each $r$-edit neighborhood until
$|\mathcal F_{s+1}|=B$ or all ranked orders have been considered.
Orders already in $\mathcal V$ or $\mathcal F_{s+1}$ are skipped.
If a neighborhood exceeds the remaining capacity, unseen orders are
admitted in enumeration order until the batch is full; the remainder
stay unevaluated and may be reconsidered in later rounds. A search batch
$\mathcal F_s$ may exceed the resident GPU pool capacity $P$ and is
processed through the reusable pool described in
Sec.~\ref{sec:gpu_solver}.


\textbf{Refinement and termination:}
Exploration stops after $s_{\max}$ expansion rounds or when no new
neighbors remain. The $K_{\mathrm{ref}}$ best feasible orders are then
re-solved with budget $t_{\mathrm{ref}}$, and the best refined feasible
trajectory is returned. Finite-budget exploration guides the discrete
search, while refinement determines the final plan.
Algorithm~\ref{alg:CoMET} summarizes the procedure.

\begin{algorithm}[t]
\caption{CoMET}
\label{alg:CoMET}
\begin{algorithmic}[1]
\Require Seed batch $\mathcal F_0$, batch width $B$, edit radius $r$, rounds $s_{\max}$, refinement count $K_{\mathrm{ref}}$, budgets $t_{\mathrm{exp}},t_{\mathrm{ref}}$, merit penalty $\mu$, feasibility tolerance $\epsilon_{\mathrm{feas}}$
\State $\mathcal V\gets\varnothing$, $s\gets0$
\While{$\mathcal F_s\neq\varnothing$ and $s\leq s_{\max}$}
    \State Evaluate $\mathcal F_s$ in the GPU pool with budget $t_{\mathrm{exp}}$ per order (Sec.~\ref{sec:gpu_solver})
    \State Store $(\widehat J,\widehat v)$ and add evaluated orders to $\mathcal V$
    \State $\mathcal F_{s+1}\gets\varnothing$
    \For{each $\mathbf m\in\mathcal V$ in increasing $\widehat M$}
        \State Add orders from $\mathcal N_r(\mathbf m)$ not in $\mathcal V\cup\mathcal F_{s+1}$, stopping at capacity $B$
        \If{$|\mathcal F_{s+1}|=B$}
            \State \textbf{break}
        \EndIf
    \EndFor
    \State $s\gets s+1$
\EndWhile
\State Re-solve the $K_{\mathrm{ref}}$ best feasible orders with budget $t_{\mathrm{ref}}$
\State \Return Best refined feasible trajectory and mode order
\end{algorithmic}
\end{algorithm}

\subsection{GPU-Parallel Mode-Conditioned TO Evaluation}
\label{sec:gpu_solver}

Candidate contact orders differ in both their active contact constraints and convergence rates, making naive lockstep GPU batching inefficient. Building on the primal--dual proximal augmented-Lagrangian DDP methods of \cite{jallet2022constrained,adeoye2025proximal}, we instead organize these heterogeneous problems as fixed-structure, independently advancing GPU solves. A shared masked representation absorbs mode-dependent constraints, fixed-workspace local systems remove dependence on the active constraint count, and a reusable resident pool accommodates heterogeneous solve times.

\paragraph{Shared representation across contact modes}
Candidates share a fixed discretization with $N$ control intervals, states $x_{1:N}$, controls $u_{0:N-1}$, and initial state $x_0$, linked by
$d_k=x_{k+1}-F_k(x_k,u_k;\mathbf m)=0$.
Each order partitions these intervals among its mode segments using a deterministic, approximately uniform allocation. When timing is optimized, interval durations are included in the controls.

Contact parameters and binary masks place mode-dependent constraints in fixed row positions. Disabled rows have zero residuals, derivatives, and multipliers. Thus, changing the contact order changes only numerical data: all candidates share the same array dimensions, memory layout, and compiled kernels within a task.

\paragraph{Fixed-workspace local solves}
Eliminating multiplier directions in the backward sweep yields local systems whose dimensions are independent of the active constraint count:
\[
S_k=H_k+\rho_k I+A_k^\top A_k/\mu_{\mathrm{AL}},
\]
where $H_k$ is local primal curvature, $A_k$ contains constraint derivatives with respect to $(u_k,x_{k+1})$, $\mu_{\mathrm{AL}}>0$ is the penalty parameter, and $\rho_k$ combines proximal regularization and adaptive damping. Masked and inactive rows contribute zeros to $A_k$.

Each system has $n_u+n_x$ unknowns, independent of the number of
active contact constraints, giving every candidate the same
factorization size and shared-memory footprint. Cholesky factorization
without pivoting checks positive definiteness; failed attempts increase
damping, and exhausted retries reject the step. A forward pass
assembles the adjustment, followed by parallel trial-step merit
evaluations. Mixed precision, internal constraint scaling, and
parameter lower bounds help control numerical conditioning.

\paragraph{Reusable pool with independent stopping}
A pool of $P$ slots retains candidate-specific trajectories,
multipliers, solver parameters, and stopping states. Each resident
candidate is mapped to a single CUDA warp, and each kernel launch
advances its primal--dual DDP solver state by one iteration.
The host periodically synchronizes, collects completed results, and
refills freed slots from the pending batch.

This avoids batch-wide synchronization on heterogeneous solve times:
unfinished candidates retain their states while completed slots are
immediately refilled from the pending batch. Search batch width $B$
determines the candidate batch size, whereas pool capacity $P$ limits
concurrent solves.

\paragraph{Stopping and scoring}
Candidates terminate upon satisfying scaled feasibility, stationarity, and dual-residual tolerances, or reaching an iteration limit. Exploration uses an iteration budget $t_{\mathrm{exp}}$. Returned trajectories are scored in the original physical units using Eq.~\eqref{eq:inexact_mode_value}, separately from the solver's internal scaling. Search feasibility requires $\widehat v\leq\epsilon_{\mathrm{feas}}$; budget exhaustion alone does not establish feasibility. Selected feasible orders are refined using the same evaluator with budget $t_{\mathrm{ref}}$.

%% file: exp.tex
\section{EXPERIMENTS}
\label{sec:exp}
We evaluate high-throughput contact-mode planning in two regimes.
Planar pushing compares against existing planners and full enumeration,
with controls separating the effects of the TO evaluator, GPU execution,
and discrete search allocation. Bimanual flipping tests search allocation
when the mode space greatly exceeds the evaluation budget, using the same
TO evaluator and budget across policies, and validates the resulting plans
on hardware.

All methods are run separately on a workstation with an NVIDIA RTX~4080 GPU, an AMD Ryzen~9 7950X3D CPU, and $30$\,GB RAM. Reported return times include host-side search, device computation, refinement, and the final trajectory re-solve.

\subsection{Experiment 1: Planar Pushing Benchmark}
\label{sec:exp-planar}

\textbf{Task and evaluation.}
Reorienting a slider requires choosing which faces to push and
when to move between them. We use the benchmark of Graesdal
et al.~\cite{Graesdal-RSS-24}: a $15$\,mm disc pusher,
quasi-static limit-surface mechanics, sticking face contact in a
$5.7^\circ$ friction cone, and the original six-term objective.
At a segment cap of eight, the eight-faced tee (Fig. \ref{fig:overview}) and four-faced box admit $36{,}465$ and
$6{,}297$ mode orders, respectively.
Each method receives the same $50$ random start and goal poses per slider.
A run succeeds if its returned trajectory satisfies its own
constraints to $\epsilon_{\mathrm{feas}}=10^{-3}$
(Eq.~\eqref{eq:inexact_mode_value}) and reaches within $5$\,mm
and $5^\circ$ of the goal. A common scorer evaluates each
method's trajectory under the benchmark objective; a separate
check, identical for every method, tests each
returned plan for physical consistency.

\textbf{CoMET configurations.}
CoMET uses $50$ knots, an optimized time step, the benchmark's
duration bound, and $t_{\mathrm{exp}}=300$ iterations per
candidate. Its pool capacity is $P=8{,}192$, expansion batch
width $B=1{,}024$, edit radius $r=2$, and refinement count
$K_{\mathrm{ref}}=256$ at $t_{\mathrm{ref}}=1{,}500$ iterations.
Each candidate starts from a schedule-derived initial guess: the pusher is placed on each scheduled face during each contact segment. Constraint rows
are scaled by their median Jacobian-row norm over sample orders;
$\widehat v$ is measured in physical units. The merit
(Eq.~\eqref{eq:search_merit}) uses $\mu=100$.
\emph{Full enumeration} evaluates every admissible order;
\emph{Seed only} evaluates only the initial batch $\mathcal{F}_0$, which contains all admissible orders with at most six segments ($2{,}289$ for the tee and $601$ for the box); and \emph{expansion}
($s_{\max}=k$) adds $k$ rounds after the seed
(Sec.~\ref{sec:parallel_enumeration}). These configurations
share the transcription, warm start, per-candidate budget,
feasibility tolerance, refinement, and scorer. Full enumeration
provides the best plan found by this procedure over the complete
order set, not a certificate of global optimality for the problems.

\textbf{Baselines and timing.}
\textbf{GCS-SDP}~\cite{Graesdal-RSS-24} runs with Mosek as
published. In the open release, its path is assigned fixed per-mode durations after
optimization, whereas CoMET optimizes timing. We therefore also
evaluate \textbf{GCS-SDP + optimal retiming}: the mode sequence
and geometric path remain fixed, and interval durations minimize
the same objective subject to the original plan's total duration.
This separable convex timing problem is solved in closed form;
retiming leaves the consistency check unchanged. The released GCS-SDP has a terminal
pusher position constraint: removing it worsened its plans by up to
$40.9\%$ in our checks. We score every method only through its
last contact, reducing the GCS-SDP cost by $6.5\%$ on the tee
and $8.4\%$ on the box. Its relaxation bounds apply to its own
formulation and do not bound CoMET's transcription. 

\textbf{IMPACT}~\cite{LiJ2-RSS-26} uses the same mechanics
model and objective, with up to three restarts.
\textbf{CMA-ES}~\cite{hansen2016cma} samples pusher
displacements at $50$ knots, using $2{,}048$ GPU-simulated
rollouts per generation.
\textbf{MCTS+IPOPT}~\cite{chen2021trajectotree,zhu2023efficient}
evaluates mode orders with IPOPT through CasADi, using MUMPS,
exact Hessians, and $16$ single-threaded processes, one per
physical core. Two CPU controls use our DDP in
\texttt{float64}: MCTS and three-round expansion, with one
candidate per thread on $16$ pinned physical cores and no BLAS
($97.8$--$98.9\%$ measured occupancy).
IMPACT, CMA-ES, and both CPU MCTS configurations have a nominal
$120$\,s cap. Seed only, full enumeration, and expansion run
to completion. GPU MCTS uses the same evaluator and refinement
as CoMET, with per-instance evaluation-count and time caps set by
three-round expansion. All MCTS variants run UCT ($c = 0.8$ by hyperparameter sweeps) with
virtual loss and random rollouts, rewarding
$\widehat J_{\mathrm{best}}/\widehat J$ for a feasible order and zero
otherwise.

\newlength{\deltaw}
\newlength{\ciw}
\settowidth{\deltaw}{\scriptsize$-10.2$}
\settowidth{\ciw}{\scriptsize$-13.1$}

\newcommand{\deltaci}[3]{%
  \makebox[\deltaw][r]{\ensuremath{#1}}%
  \ensuremath{[}%
  \makebox[\ciw][r]{\ensuremath{#2}}%
  \ensuremath{,}%
  \makebox[\ciw][r]{\ensuremath{#3}}%
  \ensuremath{]}%
}

\newcommand{\bdeltaci}[3]{%
  \makebox[\deltaw][r]{\ensuremath{\mathbf{#1}}}%
  \ensuremath{\mathbf{[}}%
  \makebox[\ciw][r]{\ensuremath{\mathbf{#2}}}%
  \ensuremath{\mathbf{,}}%
  \makebox[\ciw][r]{\ensuremath{\mathbf{#3}}}%
  \ensuremath{\mathbf{]}}%
}

\begin{table*}[t]
\vspace*{5pt}
\caption{Planar pushing ($50$ instances per slider). Wall: median return time (s); NLPs: median mode orders evaluated during exploration; $J$: median objective. Incons.: each successful plan's largest physical-consistency violation (mm), reported as median/max over plans. Paired columns compare against retimed GCS-SDP over joint successes: $\Delta J$ is the median per-instance relative objective change, brackets give bootstrap $95\%$ CIs, and Beats counts lower-cost plans. $^\dagger$GPU MCTS uses the three-round expansion budget caps. $^\ddagger$Our DDP on $16$ CPU cores.}
\label{tab:planar}

\centering
\scriptsize
\setlength{\tabcolsep}{1.8pt}
\renewcommand{\arraystretch}{1.12}

\begin{tabular}{@{}l cccc cc c c cccc cc c@{}}
\toprule
& \multicolumn{7}{c}{Push-T (8 faces, $36{,}465$ orders)}
& &
\multicolumn{7}{c}{Push-box (4 faces, $6{,}297$ orders)} \\
\cmidrule{2-8}\cmidrule{10-16}

& & & & &
\multicolumn{2}{c}{vs retimed GCS-SDP}
& & & & & & &
\multicolumn{2}{c}{vs retimed GCS-SDP}
& \\
\cmidrule{6-7}\cmidrule{14-15}

Method & Succ. & Wall & NLPs & $J$ & $\Delta J$ (\%) & Beats & Incons.
& &
Succ. & Wall & NLPs & $J$ & $\Delta J$ (\%) & Beats & Incons. \\

\midrule

GCS-SDP + optimal retiming
& 50/50 & 107.5 & --- & 22.60
& ---
& --- & $0.00/269$
& &
50/50 & 16.7 & --- & 23.75
& ---
& --- & $0.00/283$ \\

GCS-SDP~\cite{Graesdal-RSS-24}, as published
& 50/50 & 107.5 & --- & 26.67
& \deltaci{+14.0}{10.8}{18.3}
& 0/50 & $0.00/269$
& &
50/50 & 16.7 & --- & 27.57
& \deltaci{+13.8}{11.6}{16.8}
& 0/50 & $0.00/283$ \\

IMPACT~\cite{LiJ2-RSS-26}
& 15/50 & 120.0 & --- & 26.65
& \deltaci{+35.2}{-0.2}{44.5}
& 4/15 & $0.67/2.5$
& &
25/50 & 120.0 & --- & 26.68
& \deltaci{+17.6}{8.8}{39.7}
& 5/25 & $0.36/7.5$ \\

CMA-ES~\cite{hansen2016cma}
& 47/50 & 120.1 & --- & 34.01
& \deltaci{+37.6}{22.5}{57.7}
& 7/47 & $0.00/0.00$
& &
49/50 & 120.1 & --- & 31.77
& \deltaci{+16.4}{3.7}{48.8}
& 13/49 & $0.00/0.04$ \\

MCTS+IPOPT~\cite{chen2021trajectotree,zhu2023efficient}
& 50/50 & 120.0 & 324 & 23.68
& \deltaci{-0.2}{-5.1}{11.0}
& 25/50 & $0.06/0.16$
& &
50/50 & 120.0 & 402 & 23.12
& \deltaci{-6.4}{-9.4}{-2.1}
& 36/50 & $0.05/0.15$ \\

MCTS + our DDP$^\ddagger$, CPU
& 50/50 & \textbf{110.4} & \textbf{3,778} & 21.90
& \bdeltaci{-6.0}{-10.7}{-0.3}
& 32/50 & $0.09/0.51$
& &
50/50 & \textbf{108.3} & \textbf{6,182} & 22.02
& \bdeltaci{-10.2}{-13.1}{-6.0}
& 39/50 & $0.05/0.44$ \\

\midrule

\multicolumn{16}{@{}l}{\textbf{CoMET (ours) and ablations}} \\

full enumeration
& 50/50 & 113.7 & 36,465 & 21.08
& \deltaci{-6.2}{-11.2}{-2.1}
& 36/50 & $0.08/0.42$
& &
50/50 & 18.8 & 6,297 & 22.00
& \deltaci{-10.1}{-13.0}{-6.0}
& 39/50 & $0.06/0.44$ \\

expansion ($s_{\max}{=}3$)
& 50/50
& \textbf{24.6}
& \textbf{4,941}
& \textbf{21.08}
& \bdeltaci{-6.2}{-11.2}{-2.1}
& \textbf{36/50}
& $0.08/0.42$
& &
\textbf{50/50}
& \textbf{14.1}
& \textbf{3,386}
& \textbf{22.00}
& \bdeltaci{-10.1}{-13.0}{-6.0}
& \textbf{39/50}
& $0.06/0.44$ \\

expansion ($s_{\max}{=}3$)$^\ddagger$, CPU
& 50/50 & \textbf{141.3} & 4,996 & 21.12
& \deltaci{-6.1}{-11.2}{-2.0}
& 36/50 & $0.05/0.51$
& &
50/50 & \textbf{64.5} & 3,375 & 22.02
& \deltaci{-10.2}{-13.1}{-6.0}
& 39/50 & $0.05/0.44$ \\

expansion ($s_{\max}{=}1$)
& 50/50 & 17.3 & 3,313 & 21.14
& \deltaci{-6.1}{-11.2}{-1.5}
& 35/50 & $0.05/0.42$
& &
50/50 & 8.3 & 1,476 & 22.00
& \deltaci{-10.1}{-13.0}{-6.0}
& 39/50 & $0.06/0.44$ \\

MCTS$^\dagger$, $P{=}1024$
& 50/50 & 23.5 & 4,096 & 21.18
& \deltaci{-6.2}{-11.2}{-1.5}
& 36/50 & $0.06/0.42$
& &
50/50 & 12.7 & 2,048 & 22.07
& \deltaci{-10.1}{-13.0}{-6.0}
& 39/50 & $0.06/0.44$ \\

MCTS$^\dagger$, $P{=}256$
& 50/50 & 23.5 & 3,072 & 21.18
& \deltaci{-6.2}{-11.0}{-1.5}
& 35/50 & $0.05/0.42$
& &
50/50 & 12.6 & 1,792 & 22.07
& \deltaci{-10.1}{-13.0}{-6.0}
& 39/50 & $0.06/0.44$ \\

MCTS$^\dagger$, $P{=}64$
& 50/50 & 23.1 & 832 & 21.82
& \deltaci{-5.7}{-11.2}{1.3}
& 30/50 & $0.08/0.42$
& &
50/50 & 12.6 & 512 & 22.43
& \deltaci{-8.2}{-11.9}{-5.1}
& 38/50 & $0.04/1.1$ \\

seed only ($\leq6$ segments)
& 50/50 & \textbf{13.4} & 2,289 & 21.14
& \deltaci{-6.0}{-11.0}{-1.5}
& 35/50 & $0.05/0.42$
& &
50/50 & \textbf{5.6} & 601 & 22.12
& \deltaci{-8.8}{-11.4}{-5.8}
& 38/50 & $0.05/0.44$ \\

\bottomrule
\end{tabular}

\vspace{-4pt}
\end{table*}

\begin{figure}[t]
\vspace*{3pt}
\centering
\includegraphics[width=\columnwidth]{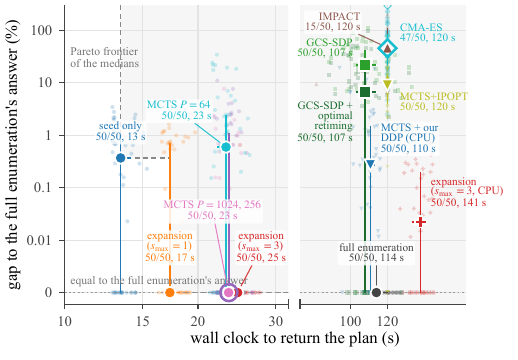}
\caption{Push-T: planning time versus objective gap to full
enumeration. Small markers show individual successful runs;
large markers summarize medians and interquartile ranges.
Open markers at the top indicate values beyond the plotted
range. The time axis has a break, and objective gaps are shown
on a logarithmic scale with a lower display limit labeled
``equal.'' Dashed lines trace the Pareto frontier of the
median results; nested markers show overlapping medians.}
\label{fig:planar-timecost}
\vspace{-3pt}
\end{figure}

\begin{figure}[t]
\vspace*{3pt}
\centering
\includegraphics[width=\columnwidth]{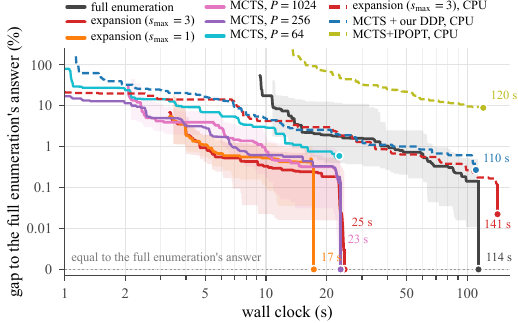}
\caption{Push-T: median gap between the best objective found
so far and the full-enumeration reference, plotted against
elapsed time. Bands show interquartile ranges, and endpoint
markers show median total planning times, including refinement.
Dashed curves use $16$ CPU threads; GPU MCTS curves use
different pool capacities $P$. }
\label{fig:planar-anytime}
\vspace{-3pt}
\end{figure}

\textbf{Solution quality and runtime.}
\textbf{CoMET} full enumeration and three-round expansion
solve every instance; expansion reduces the median return times
from $113.7$\,s and $18.8$\,s to $24.6$\,s and $14.1$\,s. Against
retimed GCS-SDP, its median paired cost
changes are $-6.2\%$ on the tee and $-10.1\%$ on the box,
with lower costs on $36/50$ and $39/50$ instances
(Table~\ref{tab:planar}, Fig.~\ref{fig:planar-timecost}).
Both paired confidence intervals lie below zero. On the tee,
expansion reduces median return time from GCS-SDP's
$107.5$\,s to $24.6$\,s, a $4.4\times$ speedup.
These results establish that broad mode evaluation
is competitive on this benchmark, while the per-problem
comparisons show that it does not dominate GCS-SDP everywhere.

We check each returned plan for five types of physical inconsistency using its motion and contact forces: penetration, force from distance, motion without force, pulling forces, and forces outside the friction cone. No method exhibits pulling forces or friction-cone violations beyond $\epsilon_{\mathrm{feas}}$.

In the tested GCS-SDP release, we observe neither penetration nor force across a gap. However, 3/50 plans for each slider rotate by $180^\circ$ within a single zero-force contact step (see supplementary video), which produces the maximum inconsistencies of 269\,mm (tee) and 283\,mm (box) in Table~\ref{tab:planar}. This is consistent with its constraint \(\sin\Delta\theta=h\omega\), which admits a \(0^\circ/180^\circ\) ambiguity. More broadly, 24/50 tee and 17/50 box plans contain a step whose rotation exceeds \(h\omega\) by over \(10^\circ\). Re-solving GCS-SDP's mode sequences with our DDP, which enforces stricter \(\Delta\theta=h\omega\), reaches feasibility on only 30/50 tee and 40/50 box instances. On the 26 tee and 33 box instances whose GCS-SDP plans are consistent with their applied wrenches, CoMET's median paired cost advantage widens to \(10.4\%\) and \(11.4\%\), with lower costs on 24/26 and 32/33 instances.

IMPACT applies pushing forces across gaps of up to $1.61$\,mm on the tee and $7.48$\,mm on the box. For \textbf{CoMET}, the measured inconsistencies remain below $0.51$\,mm, including penetration of at most $0.24$\,mm, except for GPU MCTS with $P=64$ on the box, which reaches $1.13$\,mm of penetration. IMPACT succeeds on $15/50$ tee and $25/50$ box instances, better than the results reported for the contact-implicit baseline in the GCS-SDP paper \cite{Graesdal-RSS-24}. CMA-ES succeeds more often, and both have
higher median paired costs (Table~\ref{tab:planar}). MCTS+IPOPT
succeeds throughout, with paired cost changes of $-0.2\%$
(confidence interval crossing zero) and $-6.4\%$. It evaluates
median counts of $324$ and $402$ orders within the time cap,
motivating a closer look at the evaluator's contribution.

\textbf{Evaluator and GPU contributions.}
With the same CPU resources and time cap, replacing IPOPT
with our DDP raises the median MCTS order counts from $324$
to $3{,}778$ on the tee and from $402$ to $6{,}182$ on the box.
Its paired cost changes reach $-6.0\%$ and $-10.2\%$, close to
GPU expansion's $-6.2\%$ and $-10.1\%$. Thus, much of the
quality gain on this benchmark comes from evaluating mode-conditioned
TO problems more efficiently. This comparison changes the solver
implementation and its finite-budget convergence behavior, and does not isolate evaluation breadth alone.

Running the same three-round expansion with our DDP on the CPU
produces similar costs but takes $141.3$\,s on the tee and
$64.5$\,s on the box, versus $24.6$\,s and $14.1$\,s on the GPU.
The CPU and GPU versions select the same winning order on
$33/50$ tee instances. This comparison keeps the expansion rule
and DDP algorithm, but changes execution platform and precision:
the CPU uses \texttt{float64}, whereas the GPU uses mixed precision.
Together, these controls separate two effects: the DDP evaluator
increases the useful mode evaluations attainable within a CPU budget,
while GPU-parallel execution delivers similar solution quality
substantially faster.

\textbf{How should the evaluation budget be allocated?}
Three expansion rounds recover the full-enumeration reference
on $49/50$ tee instances while evaluating a median $4{,}941$
orders, only about $14\%$ of the admissible space, reducing median
return time by $4.6\times$. On the box, expansion recovers the
reference on all $50$ instances after evaluating a median
$3{,}386$ orders, about $54\%$ of the space. The entire box
space fits within one resident pool. Seed only is less expensive but
can miss better plans: its worst tee gap is $3.5\%$, and its
box median cost is $22.12$, versus $22.00$ for full enumeration. The seed-only configuration already evaluates 2,289 mode-conditioned NLPs on the tee benchmark, highlighting the role of high-throughput evaluation even without neighborhood expansion. A check with a nine-segment space, four times larger, found no
improvement on the instances tested.

GPU MCTS exposes the trade-off between adaptation frequency and
evaluation throughput. GPU MCTS with $P=1{,}024$ recovers the full enumeration on $35/50$
tee and $45/50$ box instances under the expansion-derived
budget caps. Its completed order counts are lower than the
caps because a pool refill can be interrupted by the time budget.
Reducing $P$ to $256$ and $64$ allows more frequent search
updates, but the median tee order count falls from $4{,}096$
to $3{,}072$ and $832$, and reference recovery falls to
$31/50$ and $18/50$. Smaller pools can produce an earlier
first feasible plan (Fig.~\ref{fig:planar-anytime}), but do not
improve the returned solutions in these tests. Because $P$ changes both the update
frequency and the number of completed evaluations, these rows
reflect the computational trade-off of the current implementation
rather than the value of adaptation alone. 

Together, these ablations show that evaluator throughput is the
main performance driver on this benchmark, while the benefit of
more adaptive search is limited by its reduced evaluation throughput
in our GPU implementation.

\subsection{Experiment 2: Bimanual Flipping of an L-shaped Block}
\label{sec:exp-lflip}

\begin{figure}[t]
\vspace*{3pt}
\centering
\includegraphics[width=\columnwidth]{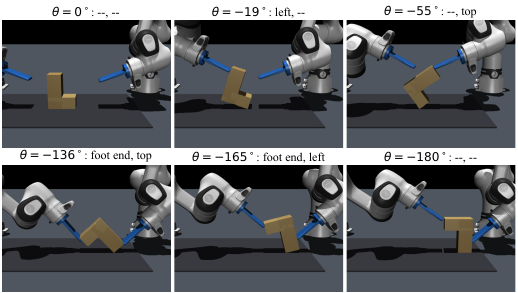}
\caption{Visualization of a planned L-block flip, with the arms
placed by inverse kinematics at the planned paddle poses.
Labels give the block angle and the two paddle contact-face
assignments; ``-{}-'' denotes free motion.}
\label{fig:lflip-strip}
\end{figure}

\begin{figure}[t]
\centering
\includegraphics[width=\columnwidth]{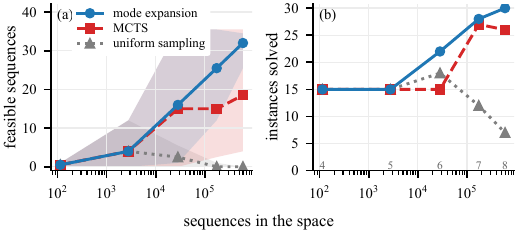}
\caption{L-flip search policies.
(a) Median feasible sequences found per run
over the \revdel{ten}\revadd{thirty} instances; bands are bootstrap $95\%$ confidence
intervals. (b) Successful instances out of \revdel{ten}\revadd{thirty} as
the segment cap increases from four to eight. Labels below
the marks give the cap; admissible-space size varies by instance.}
\label{fig:lflip-scaling}
\end{figure}

\textbf{Task and model.}
Unlike planar pushing, flipping changes the edge supporting
the object. A contact sequence must coordinate paddle motions
with these support transitions while satisfying quasi-static
wrench balance. In the planned motion, a paddle can brace the
block while the other drives a pivot, with single-paddle phases
allowing contact changes (Fig.~\ref{fig:lflip-strip}).
The L-block has a $120{\times}50$\,mm foot, a
$60{\times}150$\,mm leg, and four stable rest edges.
The transcription uses $n_x=11$, $n_u=13$, and $61$ knots,
optimizing block motion, paddle-tip poses and tilts, and
contact forces. The least-squares objective includes motion smoothness penalty, contact forces regularization, and a terminal
goal-position penalty. Mode-specific constraints place each contacting
tip on its assigned face, enforce sticking, and restrict its
force to a Coulomb cone. The table reaction is eliminated
through quasi-static wrench balance and satisfies its own
friction cone. Tool silhouettes are checked for unintended
intersections with the block, table, and each other at the
samples; continuous paddle tilts remain within an arm-feasible
band obtained from offline inverse-kinematics sweeps.

\textbf{Sequences and budget.}
An order specifies a support chain from the initial rest edge
to the goal along the convex hull in either direction.
Each paddle assigned to a face or free motion. During a pivot, at least one paddle
contacts the block; paddles cannot share a face, change faces
without a free phase, or contact the landing edge.
Under these transition rules, increasing the segment cap from four to eight expands the admissible space $\mathcal{S}$ from $94$--$134$ to $579{,}734$--$612{,}342$ orders, depending on the problem.
We test \revdel{ten}\revadd{thirty} different problems: \revdel{five}\revadd{fifteen} requiring the block to pivot successively about two support vertices and \revdel{five}\revadd{fifteen} about three, with table friction coefficients of $0.62$--$0.78$, masses of $0.10$--$0.24$ kg, and goal offsets within $\pm60$ mm.
To isolate search allocation, all policies use the same CoMET TO
evaluator, $P=512$, $t_{\mathrm{exp}}=300$, the same kinematic
warm-start construction, and at most $5{,}000$ sequence evaluations,
stopping earlier if the admissible space is exhausted.

\textbf{Our contact mode expansion} follows Algorithm~\ref{alg:CoMET},
starting from the shortest admissible sequences and expanding one-edit
neighborhoods ($r=1$) ranked by $\widehat M$ (Eq.~\eqref{eq:search_merit},
$\mu=100$).
\textbf{Uniform sampling} draws complete admissible sequences uniformly
without replacement.
\textbf{MCTS} uses UCT as in Sec.~\ref{sec:exp-planar} but rewards
$\widehat M_{\mathrm{best}}/\widehat M$ instead of the feasibility-gated
reward: feasible orders are rare here, so that reward carries little signal, and sharing the merit
gives both search policies the same information.

\textbf{Results.}
At segment caps four and five, all policies solve $15/30$ problems;
the three-pivot problems have no feasible solution in these spaces
(Fig.~\ref{fig:lflip-scaling}b). As the admissible space grows,
the policies separate. At cap eight, with approximately $600$k
orders, CoMET solves $30/30$ problems, compared with $26/30$ for
MCTS and $7/30$ for uniform sampling. CoMET also finds more feasible
orders, with a median of $32$ per run versus $18.5$ for MCTS and $0$
for uniform sampling (Fig.~\ref{fig:lflip-scaling}a).

This higher success does not come at the expense of returned-plan
quality. At caps six through eight, CoMET's returned merit is never
higher than that of the other successful policies. At cap eight,
among the $26$ problems solved by both CoMET and MCTS, the two match
on $16$, while CoMET is lower on the remaining $10$ by up to $10.7\%$.
Median runtime is similar across policies: $95.4$\,s for expansion,
$91.8$\,s for MCTS, and $91.1$\,s for uniform sampling at cap eight.

These results show that, when the mode space exceeds the evaluation
budget, local mode structure provides useful guidance for allocating
high-throughput trajectory evaluations.

\textbf{Hardware execution.}
Two Franka Emika Panda robots execute a planned three-pivot flip, tracking the planned paddle poses and tilt angles via real-time inverse kinematics (IK) at $10$\,Hz, the update rate of the three-camera AprilTag tracking system (Fig.~\ref{fig:placeholder}). The block state is used solely to verify geometric feasibility; thus, execution consists of end-effector trajectory tracking without incorporating contact dynamics. A trial is considered successful when the block comes to rest on the goal edge. Under this criterion, 12 out of 15 trials succeed (80\%). All trials are included in the supplementary video. In the three failed trials, discrepancies between the planning model and the real world cause the block to deviate from its planned trajectory, and the tracker is unable to recover. The failures highlight the limitations and suggest that force feedback and local replanning are necessary to further improve the tracking with contact.

%% file: reference.bib
@article{lozano1984automatic,
  title={{Automatic synthesis of fine-motion strategies for robots}},
  author={Lozano-Perez, Tomas and Mason, Matthew T and Taylor, Russell H},
  journal={The International Journal of Robotics Research},
  volume={3},
  number={1},
  pages={3--24},
  year={1984},
  publisher={Sage Publications Sage CA: Thousand Oaks, CA}
}

@ARTICLE{5345770,
  author={Schultz, Gerrit and Mombaur, Katja},
  journal={IEEE/ASME Transactions on Mechatronics}, 
  title={Modeling and Optimal Control of Human-Like Running}, 
  year={2010},
  volume={15},
  number={5},
  pages={783-792},
  doi={10.1109/TMECH.2009.2035112}}

@inproceedings{toussaint2015ijcai-logic,
  title     = {{Logic-Geometric Programming: An Optimization-Based Approach to Combined Task and Motion Planning}},
  author    = {Toussaint, Marc},
  booktitle = {IJCAI},
  year      = {2015},
  pages     = {1930--1936}
}

@inproceedings{cheng2022contact,
  title={{Contact mode guided motion planning for quasidynamic dexterous manipulation in 3d}},
  author={Cheng, Xianyi and Huang, Eric and Hou, Yifan and Mason, Matthew T},
  booktitle={2022 International Conference on Robotics and Automation (ICRA)},
  pages={2730--2736},
  year={2022},
  organization={IEEE}
}

@article{posa2014direct,
  title={{A direct method for trajectory optimization of rigid bodies through contact}},
  author={Posa, Michael and Cantu, Cecilia and Tedrake, Russ},
  journal={The International Journal of Robotics Research},
  volume={33},
  number={1},
  pages={69--81},
  year={2014},
  publisher={Sage Publications Sage UK: London, England}
}

@article{manchester2019contact,
  title={{Contact-implicit trajectory optimization using variational integrators}},
  author={Manchester, Zachary and Doshi, Neel and Wood, Robert J and Kuindersma, Scott},
  journal={IJRR},
  volume={38},
  number={12-13},
  pages={1463--1476},
  year={2019},
  publisher={SAGE Publications Sage UK: London, England}
}

@article{patel2019contact,
  title={{Contact-implicit trajectory optimization using orthogonal collocation}},
  author={Patel, Amir and Shield, Stacey Leigh and Kazi, Saif and Johnson, Aaron M and Biegler, Lorenz T},
  journal={IEEE Robotics and Automation Letters},
  volume={4},
  number={2},
  pages={2242--2249},
  year={2019},
  publisher={IEEE}
}

@inproceedings{onol2020tuning,
  title={{Tuning-free contact-implicit trajectory optimization}},
  author={{\"O}nol, Aykut {\"O}zgun and Corcodel, Radu and Long, Philip and Pad{\i}r, Ta{\c{s}}k{\i}n},
  booktitle={2020 IEEE International Conference on Robotics and Automation (ICRA)},
  year={2020},
}

@article{marcucci2024shortest,
  title={{Shortest paths in graphs of convex sets}},
  author={Marcucci, Tobia and Umenberger, Jack and Parrilo, Pablo and Tedrake, Russ},
  journal={SIAM Journal on Optimization},
  volume={34},
  number={1},
  pages={507--532},
  year={2024},
  publisher={SIAM}
}

@INPROCEEDINGS{Graesdal-RSS-24, 
    AUTHOR    = {Bernhard Paus Graesdal AND Shao Yuan Chew Chia AND Tobia Marcucci AND Savva Morozov AND Alexandre Amice AND Pablo Parrilo AND Russ Tedrake}, 
    TITLE     = {{Towards Tight Convex Relaxations for Contact-Rich Manipulation}}, 
    BOOKTITLE = {Proceedings of Robotics: Science and Systems}, 
    YEAR      = {2024}, 
    ADDRESS   = {Delft, Netherlands}, 
    MONTH     = {July}, 
    DOI       = {10.15607/RSS.2024.XX.132} 
}

@INPROCEEDINGS{ShenW-RSS-25, 
    AUTHOR    = {William Shen AND Caelan Reed Garrett AND Nishanth Kumar AND Ankit Goyal AND Tucker Hermans AND Leslie Pack Kaelbling AND Tomás Lozano-Pérez AND Fabio Ramos}, 
    TITLE     = {{Differentiable GPU-Parallelized Task and Motion Planning}}, 
    BOOKTITLE = {Proceedings of Robotics: Science and Systems}, 
    YEAR      = {2025}, 
    ADDRESS   = {Los Angeles, CA, USA}, 
    MONTH     = {June}, 
    DOI       = {10.15607/RSS.2025.XXI.050} 
}

@inproceedings{shen2026parallel,
  title={{Parallel Differentiable Reachability for Learning and Planning with Certified Neural Dynamics and Controllers}},
  author={Shen, Keyi and Chou, Glen},
  booktitle={Proceedings of Robotics: Science and Systems},
  year={2026}
}

@inproceedings{jallet2022constrained,
  title={{Constrained differential dynamic programming: A primal-dual augmented lagrangian approach}},
  author={Jallet, Wilson and Bambade, Antoine and Mansard, Nicolas and Carpentier, Justin},
  booktitle={2022 IEEE/RSJ International Conference on Intelligent Robots and Systems (IROS)},
  year={2022},
}

@article{adeoye2025proximal,
  title={{A proximal augmented Lagrangian method for nonconvex optimization with equality and inequality constraints}},
  author={Adeoye, Adeyemi D and Latafat, Puya and Bemporad, Alberto},
  journal={arXiv preprint arXiv:2509.02894},
  year={2025}
}

@inproceedings{natarajan2023torque,
  title={{Torque-limited manipulation planning through contact by interleaving graph search and trajectory optimization}},
  author={Natarajan, Ramkumar and Johnston, Garrison LH and Simaan, Nabil and Likhachev, Maxim and Choset, Howie},
  booktitle={2023 IEEE International Conference on Robotics and Automation (ICRA)},
  year={2023},
}

@inproceedings{toussaint2017multi,
  title={{Multi-bound tree search for logic-geometric programming in cooperative manipulation domains}},
  author={Toussaint, Marc and Lopes, Manuel},
  booktitle={2017 IEEE International Conference on Robotics and Automation (ICRA)},
  pages={4044--4051},
  year={2017},
  organization={IEEE}
}

@INPROCEEDINGS{Toussaint-RSS-18, 
    AUTHOR    = {Marc Toussaint AND Kelsey Allen AND Kevin Smith AND Joshua Tenenbaum}, 
    TITLE     = {{Differentiable Physics and Stable Modes for Tool-Use and Manipulation Planning}}, 
    BOOKTITLE = {Proceedings of RSS}, 
    YEAR      = {2018}, 
    ADDRESS   = {Pittsburgh, Pennsylvania}, 
    MONTH     = {June}, 
    DOI       = {10.15607/RSS.2018.XIV.044} 
}

@inproceedings{ortiz2022conflict,
  title={{Conflict-directed diverse planning for logic-geometric programming}},
  author={Ortiz-Haro, Joaquim and Karpas, Erez and Toussaint, Marc and Katz, Michael},
  booktitle={Proceedings of the International Conference on Automated Planning and Scheduling},
  volume={32},
  pages={279--287},
  year={2022}
}

@INPROCEEDINGS{Zhang-RSS-23, 
    AUTHOR    = {Mengchao Zhang AND Devesh K Jha AND Arvind U Raghunathan AND Kris Hauser}, 
    TITLE     = {{Simultaneous Trajectory Optimization and Contact Selection for Multi-Modal Manipulation Planning}}, 
    BOOKTITLE = {Proc. of RSS}, 
    YEAR      = {2023}, 
}

@inproceedings{chen2021trajectotree,
  title={{Trajectotree: Trajectory optimization meets tree search for planning multi-contact dexterous manipulation}},
  author={Chen, Claire and Culbertson, Preston and Lepert, Marion and Schwager, Mac and Bohg, Jeannette},
  booktitle={2021 IEEE/RSJ International Conference on Intelligent Robots and Systems (IROS)},
  pages={8262--8268},
  year={2021},
  organization={IEEE}
}

@inproceedings{sundaralingam2023curobo,
  title={{Curobo: Parallelized collision-free robot motion generation}},
  author={Sundaralingam, Balakumar and Hari, Siva Kumar Sastry and Fishman, Adam and Garrett, Caelan and Van Wyk, Karl and Blukis, Valts and Millane, Alexander and Oleynikova, Helen and Handa, Ankur and Ramos, Fabio and others},
  booktitle={2023 IEEE International Conference on Robotics and Automation (ICRA)},
  pages={8112--8119},
  year={2023},
  organization={IEEE}
}

@article{du2025gato,
  title={{GATO: GPU-accelerated and batched trajectory optimization for scalable edge model predictive control}},
  author={Du, Alexander and Adabag, Emre and Bravo-Palacios, Gabriel and Plancher, Brian},
  journal={arXiv preprint arXiv:2510.07625},
  year={2025}
}

@inproceedings{adabag2024mpcgpu,
  title={{Mpcgpu: Real-time nonlinear model predictive control through preconditioned conjugate gradient on the gpu}},
  author={Adabag, Emre and Atal, Miloni and Gerard, William and Plancher, Brian},
  booktitle={2024 IEEE International Conference on Robotics and Automation (ICRA)},
  pages={9787--9794},
  year={2024},
  organization={IEEE}
}

@inproceedings{huang2024adaptive,
  title={{Adaptive contact-implicit model predictive control with online residual learning}},
  author={Huang, Wei-Cheng and Aydinoglu, Alp and Jin, Wanxin and Posa, Michael},
  booktitle={2024 IEEE International Conference on Robotics and Automation (ICRA)},
  pages={5822--5828},
  year={2024},
  organization={IEEE}
}

@inproceedings{sleiman2019contact,
  title={{Contact-implicit trajectory optimization for dynamic object manipulation}},
  author={Sleiman, Jean-Pierre and Carius, Jan and Grandia, Ruben and Wermelinger, Martin and Hutter, Marco},
  booktitle={2019 IEEE/RSJ international conference on intelligent robots and systems (IROS)},
  pages={6814--6821},
  year={2019},
  organization={IEEE}
}

@INPROCEEDINGS{LiJ2-RSS-26, 
    AUTHOR    = {Jiayun Li AND Dejian Gong AND Georgia Chalvatzaki}, 
    TITLE     = {{IMPACT: An Implicit Active-Set Augmented Lagrangian for Fast Contact-Implicit Trajectory Optimization}}, 
    BOOKTITLE = {Proceedings of Robotics: Science and Systems}, 
    YEAR      = {2026}, 
    ADDRESS   = {Sydney, Australia}, 
    MONTH     = {July}, 
    DOI       = {10.15607/RSS.2026.XXII.163} 
}

@article{Aydinoglu2024,
  title = {{Consensus Complementarity Control for Multi-Contact MPC}},
  author = {Aydinoglu, Alp and Wei, Adam and Huang, Wei-Cheng and Posa, Michael},
  year = {2024},
  month = jul,
  journal = {IEEE Transactions on Robotics (TRO)},
  youtube = {L57Jz3dPwO8},
  arxiv = {2304.11259},
  doi = {10.1109/TRO.2024.3435423},
}

@article{le2024fast,
  title={{Fast contact-implicit model predictive control}},
  author={Le Cleac'h, Simon and Howell, Taylor A and Yang, Shuo and Lee, Chi-Yen and Zhang, John and Bishop, Arun and Schwager, Mac and Manchester, Zachary},
  journal={IEEE Transactions on Robotics},
  volume={40},
  pages={1617--1629},
  year={2024},
  publisher={IEEE}
}

@article{zhang2026sequential,
  title={{Sequential Object Placement Optimization with Convex Decomposition}},
  author={Zhang, Yuezhe and Lyu, Xiangyu and Rudra, Sohan and Tateo, Davide and Chalvatzaki, Georgia},
  journal={arXiv preprint arXiv:2608.25162},
  year={2026}
}

@INPROCEEDINGS{KangS-RSS-25, 
    AUTHOR    = {Shucheng Kang AND Guorui Liu AND Heng Yang}, 
    TITLE     = {{Global Contact-Rich Planning with Sparsity-Rich Semidefinite Relaxations}}, 
    BOOKTITLE = {Proceedings of Robotics: Science and Systems}, 
    YEAR      = {2025}, 
    ADDRESS   = {Los Angeles, CA, USA}, 
    MONTH     = {June}, 
    DOI       = {10.15607/RSS.2025.XXI.046} 
}

@article{hansen2016cma,
  title={{The {CMA} Evolution Strategy: A Tutorial}},
  author={Hansen, Nikolaus},
  journal={arXiv preprint arXiv:1604.00772},
  year={2016},
  doi={10.48550/arXiv.1604.00772}
}

@inproceedings{zhu2023efficient,
  title={{Efficient object manipulation planning with monte carlo tree search}},
  author={Zhu, Huaijiang and Meduri, Avadesh and Righetti, Ludovic},
  booktitle={2023 IEEE/RSJ IROS},
  pages={10628--10635},
  year={2023},
  organization={IEEE}
}

@article{le2025global,
  title={{Global tensor motion planning}},
  author={Le, An T and Pompetzki, Kay and Carvalho, Jo{\~a}o and Watson, Joe and Urain, Julen and Biess, Armin and Chalvatzaki, Georgia and Peters, Jan},
  journal={IEEE Robotics and Automation Letters},
  volume={10},
  number={7},
  pages={7302--7309},
  year={2025},
  publisher={IEEE}
}

@inproceedings{doshi2020hybrid,
  title={{Hybrid differential dynamic programming for planar manipulation primitives}},
  author={Doshi, Neel and Hogan, Francois R and Rodriguez, Alberto},
  booktitle={2020 IEEE International Conference on Robotics and Automation (ICRA)},
  pages={6759--6765},
  year={2020},
  organization={IEEE}
}
